\documentclass[runninheads]{llncs}
\usepackage{bm}
\usepackage[utf8]{inputenc}
\usepackage{url}
\usepackage{amssymb}
\usepackage{amsfonts}
\usepackage{graphicx}
\usepackage{url}
\usepackage{booktabs}
\usepackage{footnoterange}
\usepackage{todonotes}
\usepackage{cite}

\usepackage{hyperref}

\usepackage{doc}

\usepackage{paralist}

\title{Make E Smart Again
\\ (short paper)
\thanks{Supported by the ERC Consolidator grant no. 649043 AI4REASON
and by the Czech project AI\&Reasoning CZ.02.1.01/0.0/0.0/15\_003/
0000466 and the European Regional Development Fund.}}

\author{
Zarathustra Amadeus Goertzel
}

\institute{
  Czech Technical University in Prague, Prague
}

\authorrunning{Goertzel}
\titlerunning{Make E Smart Again}

\begin{document}

\maketitle

\begin{abstract}
In this work in progress, we demonstrate a new use-case for the ENIGMA system. The ENIGMA system using the XGBoost implementation of gradient boosted decision trees has demonstrated high capability to learn to guide the E theorem prover's inferences in real-time.
Here, we strip E to the bare bones: we replace the KBO term ordering with an identity relation as the minimal possible ordering, disable literal selection, and replace evolved strategies with a simple combination of the clause weight and FIFO (first in first out) clause evaluation functions.
We experimentally demonstrate that ENIGMA can learn to guide E as well as the smart, evolved strategies even without these standard automated theorem prover functionalities.
To this end, we experiment with XGBoost's meta-parameters over a dozen loops. \end{abstract}

\section{Introduction: Making E Stupid and Then Smart Again}
\label{sect:introduction}

State-of-the-art saturation-based automated theorem provers (ATPs) for
first-order logic (FOL), such as E and Vampire~\cite{Vampire},
employ the \emph{given clause algorithm}~\cite{Overbeek:1974:NCA:321812.321814}, translating
the input FOL problem $T\cup\{\lnot C\}$ (background theory and negated conjecture) into a refutationally
equivalent set of clauses.
The search for a contradiction is performed maintaining sets of
\emph{processed} ($P$) and \emph{unprocessed} ($U$) clauses (the \emph{proof state} $\Pi$).
The algorithm repeatedly selects a \emph{given clause} $g$ from $U$,
moves $g$ to $P$, and extends $U$ with all clauses inferred with $g$ and $P$.
This process continues until a contradiction is found, $U$ becomes empty, or
a resource limit is reached.

Historically, \emph{term ordering}, together with \emph{literal selection}, is used to guarantee the
completeness of the proof search~\cite{BG94} and to ``tame the growth of the search space and
help steer proof search'' \cite{HRSV:IJCAR-2016}.
Term ordering ensures that rewriting happens in only one direction, toward smaller terms.
Literal selection limits the inferences done with each given clause $g$ to the selected literals, which
slows down the growth of the search space and reduces redundant inferences.

E includes a \emph{strategy} language of \emph{clause evaluation functions}, made up of weight and priority functions, to heuristically guide the proof search.
In this work, I use two algorithmically invented~\cite{JakubuvU18a,JakubuvU16} strategies, E1 and E2\footnote{Strategies E1 and E2 are displayed in the appendix}, that use many sophisticated clause evaluation functions, the Knuth-Bendix ordering (KBO6), literal selection, and other E heuristics.

The ENIGMA~\cite{JakubuvU17a,JakubuvU18,JakubuvU19,GoertzelJU19} system with the XGBoost~\cite{Chen:2016:XST:2939672.2939785} implementation of gradient
boosted decision trees has recently demonstrated high capability to learn to guide the E~\cite{Schulz19} theorem prover's inferences in real-time.
ENIGMA uses the XGBoost model as a clause evaluation function to recommend clauses for selection based on clause and conjecture features.
In particular, after several proving and learning iterations, its performance on the $57880$ problems from the Mizar40~\cite{KaliszykU13b}
benchmark improved by $70\%$ ($=25397/14933$)~\cite{JakubuvU19} over the strategy E1 used for the initial proving phase.

In this work, E is stripped to the bare bones by disabling term ordering and literal selection.
KBO6 is replaced with an identity relation as the minimal possible ordering (called \texttt{IDEN} -- an addition to E
\footnote{The E version used in this paper can be found at \url{https://github.com/zariuq/eprover/tree/identity-order},
and the library for running ENIGMA with E can be found at \url{https://github.com/zariuq/enigmatic}.}).
While this frees E to do inferences in any order, E can no longer perform rewriting inferences.
The strategy E1 is replaced with the simple combination of the clause weight and FIFO (first in first out)
evaluation functions.
E is thus practically reduced to a basic superposition prover, without advanced heuristics, rewriting, or completeness guarantees. We call this strategy E0:
\begin{verbatim}
--definitional-cnf=24 --prefer-initial-clauses -tIDEN
--restrict-literal-comparisons -WNoSelection
-H'(5*Clauseweight(ConstPrio,1,1,1),1*FIFOWeight(ConstPrio))'
\end{verbatim}

E0 solves only 3872 of the Mizar40 problems in $5$ seconds compared to 14526 for E1. The first research question is the extent to which ENIGMA with this basic prover can
learn ATP guidance completely on its own.
The second is to what extent ENIGMA's learning can be boosted with data from strong
strategies and models.
That is, I explore how smart machine learning can become in this \emph{zero-strategy} setting.
The more general related question is to what extent can machine learning replace
the sophisticated human-invented theorem-proving body of wisdom used in today's ATPs for restricting advanced proof calculi.

\section{Experiments}

We evaluate ENIGMA with the basic strategy, E0, in several scenarios and over two datasets of different sizes.
All experiments are run with $5$ seconds per problem\footnote{As a rule of thumb, E solves most problems within a few seconds or \emph{not for a very long time}.}
\footnote{All the experiments are run on the same hardware unless otherwise specified: Intel(R) Xeon(R) Gold 6140 CPU @ 2.30GHz with 188GB RAM.}.

ENIGMA has so far been used in two ways: \emph{coop} combines the learned model with some standard E strategy equally ($50:50$)
while \emph{solo} only uses the learned model for choosing the given clauses.
The best results have been achieved by MaLARea-style~\cite{US+08-long} looping: that is, an ENIGMA model is trained and run with E (loop 0), then the resulting data are
added to the initial training data and a new ENIGMA model is trained (loop 1).

In this work, ENIGMA trains with both \emph{solo} and \emph{coop} data.
I present results from \emph{solo} runs because they represent the most minimal setting.

\subsection{Small Data (2000 problems)}
\begin{figure}[!h]
	\begin{centering}
	\includegraphics[scale=0.55]{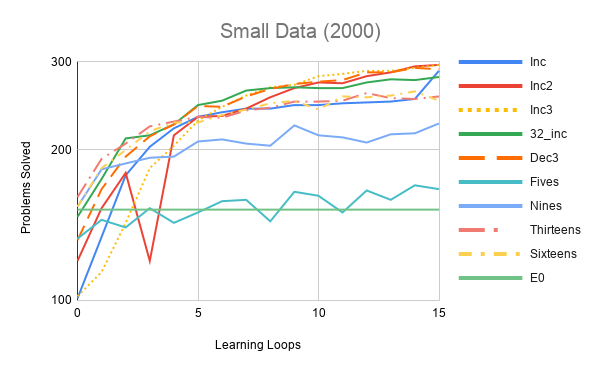}
\end{centering}
\end{figure}
The E evaluations and XGBoost training can take a long time on the full Mizar40 dataset, so
2000 randomly sampled problems are used to test meta-parameters on.
Each XGBoost model consists of $T$ decision trees of depth $D$, the most important
training meta-parameters in addition to the learning rate ($\eta=0.2$).
In previous work with ENIGMA, $T$ and $D$ were fixed for all loops of learning.
Here we try to vary the values of $T$ and $D$ during 16 loops. Let $S_{D,T}$ denote the experiment with specific $T$ and $D$.
Of the many protocols tested,
the following are included in the plot of solved problems (above): \emph{Fives} ($S_{5,100}$), \emph{Nines} ($S_{9,100}$), \emph{Thirteens} ($S_{13,200}$),
\emph{Sixteens} ($S_{16,100}$).

We also experiment with adaptively setting the meta-parameters as the
number of training examples increases according to the following protocols:

\begin{itemize}
 \item \emph{Inc} ($S_{[3,33],100}$) increases $D$ by 2 from 3 to 33 and keeps $T=100$ fixed.
 \item \emph{32\_inc} ($S_{32,[50,250]}$) fixes $D=32$ and gradually increases $T$ from 50 to 250.
 \item \emph{Inc2} ($S_{[3,33],*}$) gradually decreases $T$ from 150 to 50, varying the value intuitively
\footnote{Precise details of intuitively set parameters can be seen in the appendix.}. \item \emph{Inc3} ($S_{[3,33],[50,250]}$) aims to be more systematic and steps $T$ from 50 to 250
 \item \emph{Dec3}($S_{[3,33],[250,50]}$) decreases $T$ from 250 to 50.
\end{itemize}

At the 16th loop \emph{Inc}'s performance is best, solving 299 problems, doubling the performance of E0 (152).
However
\emph{Inc2} and \emph{Inc3} solve 298 problems and \emph{32\_inc} solves 291 problems. The conclusion is that simple protocols work well so long as $T$ or $D$ is
incremented adaptively rather than fixed.

\subsection{Big Data (57880 problems)}
\begin{figure}[!h]
	\begin{centering}
	\includegraphics[scale=0.55]{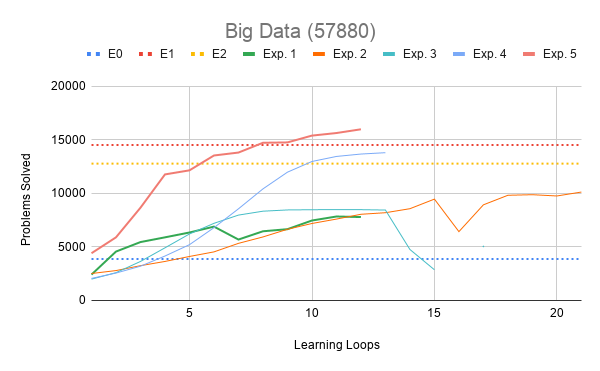}
  \label{bd}
\end{centering}
\end{figure}
These experiments are done on the large benchmark of $57880$
Mizar40~\cite{KaliszykU13b} problems from the MPTP
dataset~\cite{Urban06}.
E1 and E2 are two strong E strategies that solve 14526 and 12788 problems.

\begin{itemize}
  \item \textbf{Experiment 1} is done with $D=9$ and $T=200$ and uses our previously trained model that allowed us to
solve $25562$ problems when cooperating with E1 in our previous experiments~\cite{JakubuvU19}.
This strong model, which hashes the features into $32768$ ($2^{15}$)
buckets~\cite[Sec.~3.4]{DBLP:journals/corr/abs-1903-03182}, is used with E0 now.

\item \textbf{Experiment 2}'s parameters were intuitively toggled during the looping as in \textit{Inc3},
and a feature size of $2^{16}$ is used. Exp.~2 uses training data from E1 and E2 for additional guidance up
to the 4th loop (and then stops including them in the training data based on the
assumption they may confuse learning).
\item \textbf{Experiment 3} sets $T$ and $D$ according to protocol \emph{Inc3}.
Exp.~3 only learns from E run with E0 and trains on the GPU,
which requires the feature size to be reduced to $256$.

\item \textbf{Experiment 4} mimics Exp.~3 but uses E1 and E2 data for training (up to the 4th loop).

\item \textbf{Experiment 5} further tests boosting with data from an E0 ENIGMA
model that proved 9759 problems and an E1 that proved 21542 problems. Tree depth is
intuitively varied among 32, 512, and 1000, the number of trees is varied among
2, 100, 200, and 32. The feature vector size starts at $2^{14}$ and is decreased
to allow the data to fit on the RAM, down to 32 ($=2^{5}$).
\end{itemize}

As seen in the figure,
the strong model does not help much in guiding E without ordering or selection
in Exp.~1. Exp.~2 learns gradually and catches up with Exp.~1, but seems to
plateau around 10,000. Surprisingly the pure Exp. 3 learns fast with the
small feature size, but plateaus and drops in performance (perhaps due to overfitting).
Exp.~4 indicates that guidance is useful and surpasses E2 with 13805 in round 13.
Exp.~5 solves 15990 problems, showing that ENIGMA can take E0 beyond the smart
strategies with appropriate parameters and boosting.
This is a great improvement over the 3872 problems solved by E0.

\section{Conclusion}
ENIGMA can learn to guide the E prover effectively even without
smart strategies and term orderings. The models confer a $256\%$ increase over the naive E0 after 13 rounds of the proving/learning loop, and even trained without guidance data, a $121\%$ increase.

The experiments indicate that machine learning can be used to
fully control an ATP's guidance, learning to replace orderings,
heuristic strategies, and deal with
the increase in generated clauses without literal selection.
However the combination of ENIGMA and standard ATP heuristics still significantly out-performs ENIGMA alone.

Given the large symmetry-breaking impact of these methods in classical ATP,
future work includes, e.g., training the guidance in such a way that redundant
(symmetric) inferences are not done by the trained model once it has committed
to a certain path. This probably means equipping the learning with more history
and knowledge of the proof state in the saturation-style setting. ENIGMAWatch~\cite{GoertzelJU19}
may aid with
symmetry breaking by focusing the proof search on particular proof paths. Additional
work is needed to isolate the factors in Exp.~5's performance, and determine
the most effective boosting methods in addition to increasing $D$ and $T$ with training loops.
Ablation studies should be done to discover
the impact of term ordering and literal selection individually on E and ENIGMA's
performance. Perhaps term ordering alone is sufficient to train good ENIGMA models.

Running ENIGMA without term ordering and other restrictions is important
because it may allow us to combine training data from different strategies, and
it may allow ENIGMA to find novel proofs.

\section{Acknowledgments}
The research topic was proposed by Jan Jakubuv and Josef Urban, and further discussed with them, Martin Suda, and Thomas Tan.
I also thank the AITP'20 anonymous referees for their comments on the first extended abstract of this work.

\bibliographystyle{plain}
\bibliography{ate11,stsbib}
\appendix

\section{Strategies}

Strategy E1is:

\begin{small} \begin{verbatim}
--definitional-cnf=24 --split-aggressive --simul-paramod
--forward-context-sr --destructive-er-aggressive --destructive-er
--prefer-initial-clauses -tKBO -winvfreqrank -c1 -Ginvfreq -F1
--delete-bad-limit=150000000 -WSelectMaxLComplexAvoidPosPred
-H'(1*ConjectureTermPrefixWeight(DeferSOS,1,3,0.1,5,0,0.1,1,4),
1*ConjectureTermPrefixWeight(DeferSOS,1,3,0.5,100,0,0.2,0.2,4),
1*Refinedweight(PreferWatchlist,4,300,4,4,0.7),
1*RelevanceLevelWeight2(PreferProcessed,0,1,2,1,1,1,200,200,2.5,9999.9,9999.9),
1*StaggeredWeight(DeferSOS,1),
1*SymbolTypeweight(DeferSOS,18,7,-2,5,9999.9,2,1.5),
2*Clauseweight(PreferWatchlist,20,9999,4),
2*ConjectureSymbolWeight(DeferSOS,9999,20,50,-1,50,3,3,0.5),
2*StaggeredWeight(DeferSOS,2))'
\end{verbatim}
\end{small}

\noindent Strategy E2 is:
\begin{small} \begin{verbatim}
--definitional-cnf=24 --split-aggressive --split-reuse-defs
--simul-paramod --forward-context-sr --destructive-er-aggressive
--destructive-er --prefer-initial-clauses -tKBO -winvfreqrank
-c1 -Ginvfreq -F1 --delete-bad-limit=150000000
-WSelectMaxLComplexAvoidPosPred -H'(
3*ConjectureRelativeSymbolWeight(PreferUnitGroundGoals,0.1,100,100,50,100,0.3,1.5,1.5),
4*FIFOWeight(PreferNonGoals),
5*RelevanceLevelWeight2(ConstPrio,1,0,2,1,50,-2,-2,100,0.2,3,4))'
\end{verbatim}
\end{small}

\section{Additional Protocol Details}
In this section I include the details for \emph{intuitively toggled} protocols.

\noindent Protocol \emph{Inc2} is as follows:
\begin{center}
\begin{tabular}{ c | c c c c c c c c c c c c c c c c}
       & 0 & 1 & 2 & 3 & 4 & 5 & 6 & 7 & 8 & 9 & 10 & 11 & 12 & 13 & 14 & 15 \\ \hline
 Depth & 3 & 5 & 7 & 9 & 11 & 13 & 15 & 17 & 19 & 21 & 23 & 25 & 27 & 29 & 31 & 33 \\
 Trees & 150 & 150 & 150 & 100 & 100 & 100 & 75 & 50 & 75 & 100 & 150 & 75 & 100 & 150 & 75 & 100 \\
\end{tabular}
\end{center}

\noindent The protocol for Exp.~2 is as follows:
\begin{center}
  \begin{footnotesize}
   \setlength\tabcolsep{1pt}
\begin{tabular}{ c | c c c c c c c c c c c c c c c c c c c c c c}
       & 0 & 1 & 2 & 3 & 4 & 5 & 6 & 7 & 8 & 9 & 10 & 11 & 12 & 13 & 14 & 15 & 16 & 17 & 18 & 19 & 20 & 21 \\ \hline
 D & 4 & 5 & 6 & 7 & 8 & 9 & 10 & 11 & 12 & 13 & 14 & 15 & 16 & 16 & 32 & 9 & 16 & 32 & 64 & 24 & 25 & 32 \\
 T & 50 & 150 & 160 & 170 & 180 & 190 & 200 & 200 & 200 & 200 & 210 & 220 & 225 & 225 & 225 & 300 & 300 & 225 & 150 & 250 & 250 & 250 \\
\end{tabular}
  \end{footnotesize}
\end{center}

\noindent The protocol for Exp~.5 requires some explanation.
The motivation is to see how far E0 can be taken, even if the methods are too CPU-intensive for a thorough grid search.

Exp.~2 and Exp.~4 demonstrate the utility of boosting. Thus to create better boosting data I trained ENIGMA for 10 loops with strategies E1 through E12 and used this as boosting data for the first 4 of 10 loops of training.
In addition to training E0, and in the spirit of ablation studies, I also trained ENIGMA models for E0 with KBO ordering (and no literal selection) and for E0 with KBO ordering and restricted literal comparisons. The motivation is that these versions may serve as a bridge between standard E and the basic E0.

Then I used these results to boost an ENIGMA model in loop 0, and trained based on this for 10 loops, proving 9759 problems.

Finally this data and the data from a loop 3 ENIGMA model trained with E1 is used to boost E0 with the following meta-parameters:
\begin{center}
\begin{tabular}{ c | c c c c c c c c c c c c c c c c}
       & 0 & 1 & 2 & 3 & 4 & 5 & 6 & 7 & 8 & 9 & 10 & 11 \\ \hline
 Depth & 512 & 512 & 32 & 1000 & 32 & 1000 & 32 & 1000 & 32 & 1000 & 1000 & 100 \\
 Trees & 2 & 2 & 100 & 100 & 200 & 100 & 200 & 32 & 300 & 32 & 32 & 32 \\
 Feature Size & 16384 & 8192 & 4096 & 28 & 4096 & 28 & 4096 & 32 & 2048 & 64 & 32 & 128 \\
\end{tabular}
\end{center}

\end{document}